\documentclass{article}

    \PassOptionsToPackage{numbers, compress}{natbib}

    \usepackage[preprint]{neurips_2023}

\usepackage[utf8]{inputenc} 
\usepackage[T1]{fontenc}    
\usepackage{hyperref}       
\usepackage{url}            
\usepackage{booktabs}       
\usepackage{amsfonts}       
\usepackage{nicefrac}       
\usepackage{microtype}      
\usepackage[table]{xcolor}         

\usepackage[ruled,vlined]{algorithm2e}
\usepackage{amsmath}
\usepackage{multirow}
\usepackage{diagbox}
\usepackage{graphicx}
\usepackage{subfigure}
\usepackage{makecell}
\newcommand{\ours}{\mbox{E-CGL}}
\newcommand{\eg}{\textit{e.g.}}
\newcommand{\ie}{\textit{i.e.}}
\newcommand{\etc}{\textit{etc.}}

\title{E-CGL: An Efficient Continual Graph Learner}

\author{%
  Jianhao Guo \\
  Zhejiang University \\
  \texttt{guojianhao@zju.edu.cn} \\
  \And
  Zixuan Ni \\
  Zhejiang University \\
  \texttt{zixuan2i@zju.edu.cn} \\
  \And
  Yun Zhu \\
  Zhejiang University \\
  \texttt{zhuyun\_dcd@zju.edu.cn} \\
  \And
  Siliang Tang \\
  Zhejiang University \\
  \texttt{siliang@zju.edu.cn} \\
}

\begin{document}

\maketitle
\begin{abstract}
Continual learning has emerged as a crucial paradigm for learning from sequential data while preserving previous knowledge. 
In the realm of continual graph learning, where graphs continuously evolve based on streaming graph data, continual graph learning presents unique challenges that require adaptive and efficient graph learning methods in addition to the problem of catastrophic forgetting.
The first challenge arises from the \textbf{interdependencies} between different graph data, where previous graphs can influence new data distributions. 
The second challenge lies in the \textbf{efficiency} concern when dealing with large graphs. 
To addresses these two problems, we produce an Efficient Continual Graph Learner (E-CGL) in this paper. 
We tackle the \textbf{interdependencies} issue by demonstrating the effectiveness of replay strategies and introducing a combined sampling strategy that considers both node importance and diversity.
To overcome the limitation of \textbf{efficiency}, E-CGL leverages a simple yet effective MLP model that shares weights with a GCN during training, achieving acceleration by circumventing the computationally expensive message passing process.  
Our method comprehensively surpasses nine baselines on four graph continual learning datasets under two settings, meanwhile E-CGL largely reduces the catastrophic forgetting problem down to an average of -1.1\%.
Additionally, E-CGL achieves an average of 15.83$\times$ training time acceleration and 4.89$\times$ inference time acceleration across the four datasets. These results indicate that E-CGL not only effectively manages the correlation between different graph data during continual training but also enhances the efficiency of continual learning on large graphs.
The code is publicly available at \url{https://github.com/aubreygjh/E-CGL}.

\end{abstract}
\section{Introduction}
\label{sec:intro}

Graphs have garnered significant research attention in recent years due to their ubiquity as a data form~\cite{gcn,gat,gin}. In real-world applications, graph data tends to expand and new patterns emerge over time. For instance, recommendation networks may introduce new product categories, citation networks may witness the emergence of new types of papers and research hotspots, and chemical design may uncover novel molecules and drugs. To keep up with these evolving scenarios and provide up-to-date predictions, graph models need to continuously adapt. 
However, traditional training strategies~\cite{niepert2016,gcn} will suffer from catastrophic forgetting~\cite{yuan2023continual} while adapting the model on new data, resulting in poor performance in previous tasks.

\begin{figure}[htb]
  \centering
  \includegraphics[scale=0.5]{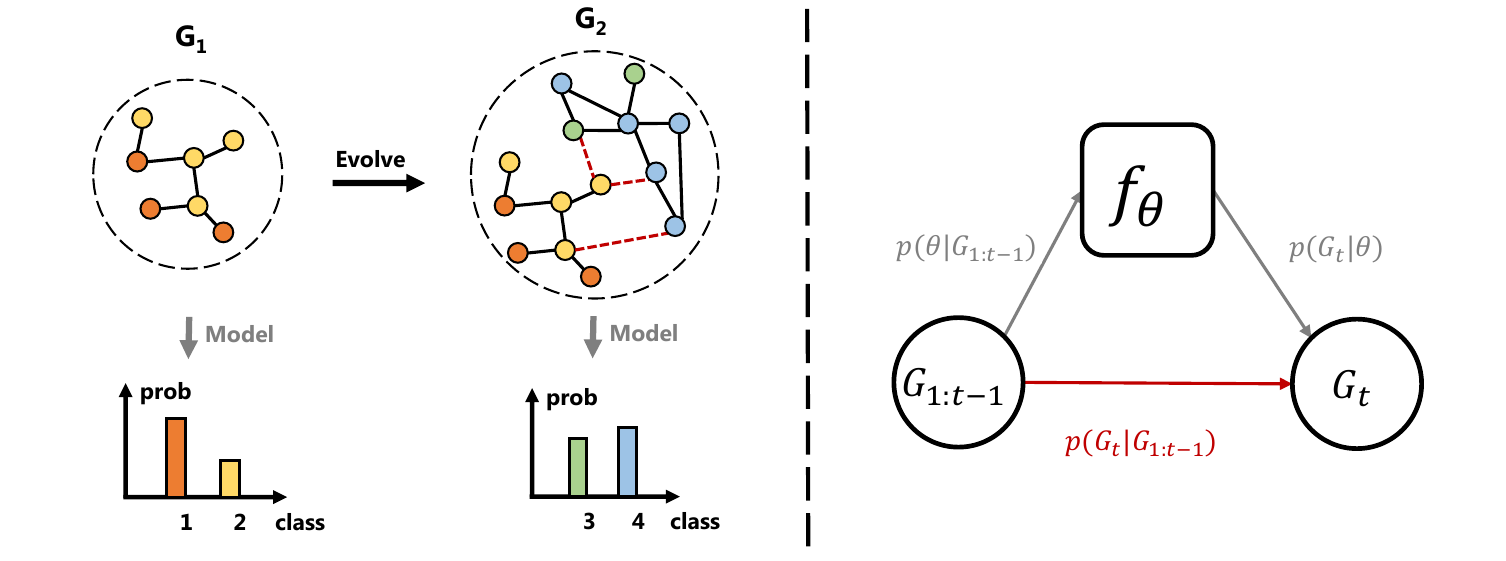}
  \label{fig:1}
  \caption{Left: Visualization of continual graph learning.
Right: Illustration of conditional probabilities on continual graph learning. The grey lines show Bayes rule for independent identically distributed data. The red line represents the influence of previous data on the current graph.}
\end{figure}

Hence, there is a need for methods that can rapidly adapt to new classes while maintaining performance on previous tasks. Continual learning (also known as incremental learning or lifelong learning) aims to achieve this by learning from new data while preserving previous knowledge~\cite{thrun1995lifelong}. Continual learning has been extensively studied in areas such as computer vision ~\cite{rebuffi2017icarl,buzzega2020dark,ni22modx} and natural language processing~\cite{srinivasan2022climb,fan2022unified}, among others. However, applying continual learning to graphs presents unique challenges.

\textbf{The first challenge arises from the interdependent nature of graph data, wherein new data distribution is conditioned on the old data.}
We illustrate this challenge in Figure \ref{fig:1}. Under the independent  setup like images (shown in grey lines), the network $f_\theta$ absorbs information learned on previous data with $p(\theta|G_{1:t-1})$ and expects to fit on current data with $p(G_t|\theta)$, here $f_\theta$ prevents previous data from explicitly influencing current distribution.
But on graphs, due to the existence of inter-task edges (shown in red line), the likelihood for current data is also conditioned on previous data by $p(G_t|G_{1:t-1})$.
Thus we have $p(\theta|G_{1:t}) \propto \frac{p(\textcolor{red}{G_t|G_{1:t-1}}, \theta)p(\theta|G_{1:t-1})}{p(G_t)}$, where the red term indicates the dependency between old and new graphs.
This would add more difficulty for continual graph learning since previous data will not be fully available in this scenario. 
\textbf{The second challenge refers to efficiency concern on increasing large graphs.}
In real-world scenarios such as social networks and recommendation systems, it is crucial to quickly capture emerging trends to enhance user experience and product competitiveness. However, as the number of nodes and edges in graph data increases, the computational cost grows exponentially, making efficient model updates increasingly challenging.

Recently, there has been a rapid growth in research interest regarding continual graph learning due to its practical applications. Several pioneering works have extended the ideas of classical continual learning to graphs ~\cite{yuan2023continual,febrinanto2023graph}. 
Representative works include TWP~\cite{twp}, which adopts parameter regularization and constrains the drastic shift of parameter space, and ER-GNN~\cite{ergnn}, which proposes intuitive node selection strategies for replaying, \etc~
However, they do not adequately address the unique challenges posed by graph dependencies, nor do they fully exploit graph topological and attributive information.
Additionally, these works primarily focus on addressing catastrophic forgetting and overlook the efficiency problem in practical applications involving large-scale graphs. 

In this paper, we propose E-CGL, an \textbf{E}fficient \textbf{C}ontinual \textbf{G}raph \textbf{L}earner, to tackle the aforementioned challenges in continual graph learning. 
\textbf{To model the dependencies between tasks and alleviate catastrophic forgetting}, we demonstrate the effectiveness of the replay strategy for continual graph learning and propose a combined sampling strategy that considers both node importance and diversity. Firstly, we design an importance sampling strategy that incorporates both topology and attribution aspects by leveraging an attributed-PageRank algorithm. Secondly, we introduce a diversity sampling strategy that captures novel patterns by modeling the differences between a node and its neighbors.
\textbf{To improve training efficiency}, we utilize a simple yet effective MLP model that shares the same weight space with its counterpart GCN during training. This eliminates the time-consuming message passing process. During inference, we leverage the graph structure information to enhance prediction performance.
Our method achieves the best results in both average accuracy and average forgetting across four large-scale graph continual learning datasets under two challenging settings, effectively improving model performance while alleviating catastrophic forgetting. 
Furthermore, in comparison to GCN methods, E-CGL exhibits an average training/inference speed increase of 15.83/4.89 times, with a maximum increase of 28.44/8.89 training/inference speed on the OGBN-Products dataset.

The contribution of this work can be summarized as follows:
\textbf{(1)} We identify two novel challenges when applying continual learning to graphs: the interdependencies of graph data, which can be influenced by previous knowledge; and the training efficiency on large growing graphs.
\textbf{(2)} To address these challenges, we propose an efficient continual graph learner that employs graph dependent sampling strategies to recover interdependencies and an efficient MLP encoder to accelerate the training process.
\textbf{(3)} Empirical evaluations on four large node classification datasets demonstrate the state-of-the-art performance and validate the effectiveness of E-CGL. 

\section{Related Works}
Continual learning (CL) aims at progressively acquiring and integrating knowledge from new task without forgetting previously learned information.
A naive solution to continual learning is retraining the model with both old and new data, however, this is impractical since it is time-consuming, labor-intensive, and costly for real-world applications.
Another solution involves fine-tuning the model with only new data, but this often leads to catastrophic forgetting, where the model rapidly declines in its ability to distinguish old classes. Hence, both training efficiency and robustness to catastrophic forgetting are crucial for successful continual learning.

Existing approaches for continual learning can be roughly divided into three categories: regularization-based, replay-based, and architecture-based.
Regularization-based methods penalize changes in the model's parameters by incorporating regularization terms to preserve past task performance. For instance, LwF leverages knowledge distillation to regularize the model's parameters, it uses the previous model's output as soft labels for current tasks. EWC~\cite{ewc}, on the other hand, directly adds quadratic penalties to the model weights to prevent drastic parameter shifting. MAS~\cite{mas} is similar to EWC, that uses the sensitivity of predictions on parameters as the regularization term.
Replay-based methods prevent forgetting by selecting a set of exemplary data from prior tasks to retrain the model along with new data. For example, GEM~\cite{gem} stores representative data in the episodic memory and modifies the gradients of the current task to prevent forgetting.
Architecture-based approaches introduce different parameter subsets for different tasks to avoid drastic changes.
While most of these methods can be integrated into graph learning with proper modifications, they often overlook the distinct properties of graph data and result in performance degradation.

Recently, with a growing interest in applying continual learning to graph data, several specific methods have been proposed. 
TWP~\cite{twp} adds a penalty to preserve the topological information of the previous graphs. 
ER-GNN~\cite{ergnn} integrates three intuitive replay strategies to GNNs. However, these node sampling techniques are tailored for i.i.d. data, and do not take graph topology into consideration. 
ContinualGNN~\cite{continualgnn} is another replay-based approach that samples nodes based on novel pattern discovery and old pattern preservation. 
DyGRAIN~\cite{dygrain} proposes time-varying receptive fields to capture pattern changes.
SSM~\cite{ssm} sparsifies the replayed graph to reduce memory consumption.
CaT~\cite{cat} employs graph condensation techniques to generate synthesized replayed graphs.
Furthermore, CGLB~\cite{cglb} summarizes and proposes unified benchmark tasks and evaluation protocols for continual graph learning, highlighting the importance of inter-task and intra-task edges among different tasks.

Despite the promising prospects, the field of Continual Graph Learning (CGL) still lacks sufficient research contributions, and several challenges remain to be addressed. In addition to the interdependency and efficiency problems that we aim to improve in our work, there are other open issues that require attention. These include investigating the impact of neighborhood evolution on continual learning, addressing the complexities of continual learning on heterogeneous graphs, devising strategies for actively forgetting stale patterns, and more. Further research in the field of CGL is essential to tackle these challenges and advance our understanding of continual learning on graphs.
\section{Methodology}
In this section, we will delve into the technical details of E-CGL. Firstly, we will provide the fundamental concepts of continual graph learning. Following that, we will present two essential components of our approach, namely Graph Dependent Replay and Efficient Graph Learner. Lastly, we will conduct a complexity analysis to demonstrate the efficiency of E-CGL.

\subsection{Preliminaries}
\paragraph{Notations} A growing graph can be decomposed into a series of dynamic graphs $\mathcal{G}= \{G_1, G_2,\dots,G_T\}$ according to timestamps. 
Each subgraph $G_t = \left(\mathcal{V}_t,\mathcal{E}_t\right)$ constitutes a distinct task $t$ and has different types of nodes $\mathcal{V}_t$ and edges $\mathcal{E}_t$.
In the context of node classification tasks in a continual learning setting, we denote $\mathbf{A}_t \in \mathbb{R}^{N_t \times N_t}$ as the adjacency matrix, $\mathbf{X}_t \in \mathbb{R}^{N_t\times K}$ as the raw node features, and $\mathbf{Y}_t \in \mathbb{R}^{N_t\times C}$ as the one-hot node labels for $G_t$. Here, $N_t = |\mathcal{V}_t|$ represents the number of nodes, $K$ denotes the dimension of raw features, and $C$ represents the total number of classes. 
$\mathcal{M}$ is a memory bank that stores historical data.
Since the methodologies in this section are all based on the current task $t$, unless otherwise specified, we will omit the subscript $t$ for brevity.

\paragraph{Problem Definition}
In the continual graph learning setting, each subgraph $G_t$ in $\mathcal{G}$ has no overlap in category labels, and only the data at current time $t$ is visible to the model due to the storage limitation. The goal is to learn the newly emerging information while prevent catastrophic forgetting of previous knowledge. 
For each subgraph $G_t$, we split it into training set $G_t^{tr}=\left(\mathcal{V}^{tr}_t,\mathcal{E}^{tr}_t\right)$ and testing set $G_t^{te}=\left(\mathcal{V}^{te}_t,\mathcal{E}^{te}_t\right)$ to train and evaluate the current model $f$ parameterized by $\theta_t$.

\subsection{Graph Dependent Replay}
As mentioned in Section \ref{sec:intro}, continual graph learning faces novel challenges for alleviating catastrophic forgetting due to the topological dependencies of nodes and graphs. To overcome these challenges, a direct solution is to replay historical data and rebuild such dependencies.
And to justify replay-based strategy from a probabilistic perspective, we compute the posterior probability $p(\theta|\mathcal{D})$ by applying Bayes's rule to the prior probability of the parameters $p(\theta)$ and likelihood function $p(\mathcal{D}|\theta)$:
\begin{equation}
\log p(\theta|\mathcal{D}) = \log p(\mathcal{D}|\theta) + \log p(\theta) - \log p(\mathcal{D}),
\label{eq:probalility_optimize}
\end{equation}
where $\mathcal{D}$ represents the dataset. By rearranging Equation \ref{eq:probalility_optimize} and considering the combination of new data $\mathcal{D}_{new}$ and previous data $\mathcal{D}_{old}$, we have:
\begin{equation}
\begin{split}
\log p(\theta|\mathcal{D}_{new} \cup \mathcal{D}_{old}) = & \log p(\mathcal{D}_{new}|\theta, \mathcal{D}_{old}) + \log p(\theta|\mathcal{D}_{old}) \\ 
&-\log p(\mathcal{D}_{new}).
\end{split}
\label{eq:continual_probalility_optimize}
\end{equation}
The first term $\log p(\mathcal{D}_{new}|\theta, \mathcal{D}_{old})$ on the right-hand side shows the interdependencies of graph data, where the distribution of new data can be influenced by the old data.
Such interdependencies make the problem challenging to address if the original data $\mathcal{D}_{old}$ is unavailable. 
Based on above analysis, we opt to use replay-based methods to prevent catastrophic forgetting under the continual graph learning scenario.
In general, the replay-based strategy maintains a memory buffer $\mathcal{M}$, which is used to store historical data from previous tasks. When a new task $t$ arrives, the model learns from the new data, and replays from the old data:
\begin{align}
    \mathcal{L}_{\text{new}}=\sum_{i \in \mathcal{V}^{tr} }L_\text{CE}(f(\mathbf{x}_{i}), \mathbf{y}_{i}),
    \quad
    \mathcal{L}_{\text{replay}}=\sum_{j \in \mathcal{M}}L_\text{CE}(f(\mathbf{x}_{j}), \mathbf{y}_{j}),
\label{eq:loss_1}
\end{align}
where $L_\text{CE}(f(\mathbf{x}_{i}), \mathbf{y}_{i})=-\mathbf{y}_i \log f(\mathbf{x}_i)$ refers to the cross-entropy loss.
Due to storage limit, it is impractical to replay all historical data, and effective sampling strategies are needed. In this work, we present a novel sampling strategy that takes into account both the \textit{importance} and \textit{diversity} of graph data, utilizing their topological and attributive characteristics.

\paragraph{Importance Sampling}
PageRank was proposed by Google~\cite{pagerank} as a web page ranking algorithm. It evaluates the importance of each node by considering the importance of other nodes linked to it. 
Formally, it iteratively updates the ranking scores using the 1st-order Markov chain until convergence:
\begin{align}
    \pi = dT\pi+\frac{1-d}{N}\mathbf{1},
    \quad
    T_{i j} = 
                \begin{cases}
                \frac{1}{\delta_j} & \text { if directed edge }(j, i) \in \mathcal{E} \\ 
                \frac{1}{N} & \text { if } \delta_j=0 \\ 
                0 & \text { otherwise }
                \end{cases},
\end{align}
where $\pi\in\mathbb{R}^{N \times 1}$ denotes the PageRank scores for $N$ nodes, $d$ is the damping factor, $T$ is the transition matrix based on graph topology, $\delta_j$ is the out-degree of node $j$, and $\mathbf{1}$ is a column vector valued as 1. Under the Markov chain framework, the ranking scores $\pi$ can be interpreted as the transition probability distribution of a random walk on the graph.
Nevertheless, a major defect of PageRank is that it only considers the topological structures of a graph, while neglecting the node attributive information. This could be problematic when dealing with large scale graphs with rich attributes. Inspired by the work of ~\cite{attrirank}, we propose to utilize the node attributes as well as the graph topology for importance ranking.

Consider the homophilous assumption~\cite{h2gcn}, that nodes with similar attributes tend to share similar labels and importance. 
Hence, we employ another random walk based on attribute similarity matrix as transition probability. Defining a fully connected version of current graph $G^{fc}$, whose edge weight is the similarity of its connected nodes $s(i,j)$, then the transition matrix of $G^{fc}$ is given as:
\begin{equation}
    Q_{ij}=\frac{s(i,j)}{\sum_{k \in \mathcal{V}}s(k,j)},
\label{eq:Q}
\end{equation}
wherein we use RBF kernel function~\cite{rbf} as similarity metric: $s(i,j)=e^{-\gamma||\mathbf{x}_i-\mathbf{x}_j||_2^2}$. In this way, the value of $Q_{i,j}$ is guaranteed in $[0,1]$.
With the attribute-based transition matrix, combining the topology-based transition matrix of vanilla PageRank, the updating strategy for importance ranking is derived as:
\begin{equation}
    \pi_{\text{Imp}} = dT\pi_{\text{Imp}}+(1-d)Q\pi_{\text{Imp}},
\label{eq:attrirank}
\end{equation}
where $d\in [0,1]$ serves as the controlling parameter that balances the ratio of $T$ and $Q$.
However, the computation of $Q$ and update of $Q\pi_{\text{Imp}}$ takes the time complexity of $O(|\mathcal{V}|^2)$, which is unacceptable for large graphs. To solve the issue, we use an approximation trick~\cite{attrirank} to simplify the process.
First we define a vector as:
\begin{equation}
    r_i = \frac{1}{z}\sum_{j \in \mathcal{V}}s(i,j),
\label{eq:r}
\end{equation}
where $z=\sum_i\sum_js(i,j)$ is the normalization term. From the definition we obtain:
\begin{equation}
    1\cdot r = Qr.
\label{eq:surrogate}
\end{equation}

The proof for Equation \ref{eq:surrogate} is in Appendix \ref{appendix:proof1}. From this equation we find the vector $r$ serves as the stationary probability distribution for $Q$, as well as the corresponding eigenvector of the largest eigenvalue 1 of $Q$. Therefore, we use $r$ as the surrogate of $Q\pi_{\text{Imp}}$, and Equation \ref{eq:attrirank} is reformed as:
\begin{equation}
    \pi_{\text{Imp}} = dT\pi_{\text{Imp}}+(1-d)r.
\label{eq:attrirank_simple}
\end{equation}

With $r$ to replace $Q\pi_{\text{Imp}}$ and RBF as similarity, the importance ranking complexity in Equation \ref{eq:attrirank_simple} can be further reduced to $O(|\mathcal{V}|K^2)$ by Taylor expansion $e^x \approx 1+x+\frac{1}{2}x^2 (x \rightarrow 0)$. We defer the detailed derivation in Appendix \ref{appendix:proof2}.

\paragraph{Diversity Sampling}
The algorithm mentioned above selects important nodes to counteract catastrophic forgetting. However, it is equally important to consider instances that are vulnerable to subsequent tasks. The influence on the old graph could lead to the emergence of new patterns while diminishing old ones. To capture these patterns, we propose a heuristic strategy that samples nodes to enhance the diversity of the memory buffer.

Specifically, we define nodes as diverse if they significantly differ from their surrounding nodes at the feature level. To determine diversity, we calculate the average feature of a node's neighbors (using only 1-hop neighbors for efficiency), and then compare it with the feature of the node itself:
\begin{equation}
    \pi_{\text{Div}}(i)=||\mathbf{x}_i -\frac{1}{|\mathcal{N}_i|}\sum_{u\in\mathcal{N}_i}\mathbf{x}_u||.
\label{eq:divrank}
\end{equation}

$\pi_{\text{Div}}(i)$ reflects the diversity between node $i$ and its surrounding environment, and nodes with higher diversity may be vulnerable to new patterns and should be replayed in the future.
Overall, the memory buffer $\mathcal{M}$ is updated by :
\begin{align}
\mathcal{M} = \mathcal{M} \cup \text{argtopk}\pi_{\text{Imp}} \cup \text{argtopk}\pi_{\text{Div}}.  
\end{align}

\subsection{Efficient Graph Learner}
To address the efficiency challenge in practical applications involving large growing graphs, we propose an efficient graph learner. This approach involves incorporating a Multi-Layer Perceptron (MLP) for training purpose, and utilizing its weights to initialize the GNN for inference tasks.
Specifically, a typical graph convolution network~\cite{gcn} layer can be formulated as:
\begin{align}
    \mathbf{H}^{(l)}=\sigma\left(\mathbf{\tilde{A}}\mathbf{H}^{(l-1)}\mathbf{W}^{(l)}_\text{GCN}\right),
\label{eq:gcn}
\end{align}
where $\sigma$ is the non-linear activation function,  $\mathbf{\tilde{A}}$ is the normalized adjacency matrix~\cite{gcn}, $\mathbf{H}^{(0)}=\mathbf{X}$ is the input node features, and $\mathbf{W}^{(l)}_\text{GCN}$ are the learnable weight matrix of the $l$-th layer. We can further decouple the formulation into two operations, \ie, message passing and feature transformation~\cite{sgc}:
\begin{align}
 \textit{message passing: }& \hat{\mathbf{H}}^{(l-1)}=\mathbf{\tilde{A}}\mathbf{H}^{(l-1)}, \\
 \quad \textit{feature transformation: }& \mathbf{H}^{(l)}=\sigma\left(\hat{\mathbf{H}}^{(l-1)}\mathbf{W}^{(l)}_\text{GCN}\right).
\end{align}

Message passing mechanism, which works by aggregating information from each node's neighborhood, has long been considered as the core part for GNN models (\eg, GCN~\cite{gcn}, GraphSAGE~\cite{graphsage}, GAT~\cite{gat}). 
However, it also consumes most of computation time due to the sparse matrix multiplication.
Recent studies~\cite{pmlp,mlpinit} have found that the effect of message passing mainly come from its generalization ability in inference, rather than its representation ability of learning better features in training.
This finding opens up the possibility of removing message passing during training to improve efficiency.
Besides, message passing is non-parametric in most cases, which means it can be plug-and-play integrated into existing models without training.
Consequently, Equation \ref{eq:gcn} degrades to a simple MLP layer by removing message passing:
\begin{align}
    \mathbf{H}^{(l)}=\sigma\left(\mathbf{H}^{(l-1)}\mathbf{W}^{(l)}_\text{MLP}\right).
\end{align}
Note that when the dimensions of hidden layers are set same for $\mathbf{W}_\text{MLP}$ and $\mathbf{W}_\text{GCN}$, they will have identical weight space, meaning the weights of an MLP and its counterpart GCN can be transferred to each other. 
Based on such premise, we propose an efficient graph learner for fast model adaptation under continual setting. 
Concretely, we first remove the message passing scheme of a GCN network and initialize its counterpart MLP network. Then we optimize the parameters $\mathbf{W}_\text{MLP}$ with only the node features as input, which speeds up the training process by avoiding sparse matrix multiplication.
Formally, the training objective of Equation \ref{eq:loss_1} with MLP is derived as:
\begin{align}
    \mathcal{L}_{\text{new}}&=\sum_{i \in \mathcal{V}^{tr} }L_\text{CE}(f_{\text{MLP}}(\mathbf{x}_{i};\mathbf{W}_{\text{MLP}}), \mathbf{y}_{i}), \\
    \quad
    \mathcal{L}_{\text{replay}}&=\sum_{j \in \mathcal{M}}L_\text{CE}(f_{\text{MLP}}(\mathbf{x}_{j};\mathbf{W}_{\text{MLP}}), \mathbf{y}_{j}),
\label{eq:loss_2}
\end{align}

During inference, we adopt the trained $\mathbf{W}_\text{MLP}$ as the parameters of the corresponding GCN model, and the topological information is leveraged again with message passing added:
\begin{align}
    \mathbf{H}^{(l)}&=\sigma\left(\mathbf{\tilde{A}}\mathbf{H}^{(l-1)}\mathbf{W}^{(l)}_\text{MLP}\right),\\
    \hat{\mathbf{Y}}&=\text{softmax}(\mathbf{H}^{(L)}).
\end{align}

Note that the GCN encoder can be generalized to any message-passing-based GNN networks with proper modifications, we use GCN here for notation simplicity. 
Empirical results show that our efficient graph learner achieves comparable or even better results compared with GNNs, but largely reduce the training time up to 15.83$\times$.

\subsection{Discussions}

\paragraph{Training Objective}
Integrating both Graph Dependent Replay module and Efficient Graph Learner module, the framework of our E-CGL is outlined in Algorithm \ref{alg:1}.
The overall training objective for E-CGL at each task is formulated as:
\begin{align}
    \mathcal{L} = \mathcal{L}_\text{new}+\lambda\mathcal{L}_\text{replay},
\end{align}
where $\lambda$ is a weight that balances the strength of two losses, we simply set it as 1.

\begin{algorithm}[ht]
\SetNoFillComment
\SetAlgoLined
\caption{Framework of E-CGL}
\label{alg:1}
\KwIn{Continual graphs: $\mathcal{G}= \{G_1, G_2,\dots,G_T\}$; Memory bank: $\mathcal{M}$;  max epochs $E$}
\KwOut{Classification model $f$ parameterized by $\mathbf{W}$}
\BlankLine
\For{$t \leftarrow 1$ \KwTo $T$}{
\tcc{Train} 

Obtain current training set $G_t=(\mathcal{V}_{t}^{tr},\mathcal{E}_{t}^{tr})$ and memory buffer $\mathcal{M}$ 

\For{$epoch \leftarrow 1$ \KwTo $E$}{
    \tcp{Compute loss function:}
    
    $\mathcal{L}_{\text{new}}=\sum_{i \in \mathcal{V}_{t}^{tr} }L_\text{CE}(f_{\text{MLP}}(\mathbf{x}_{i};\mathbf{W}_{\text{MLP}}), \mathbf{y}_{i})$
    
    $\mathcal{L}_{\text{replay}}=\sum_{j \in \mathcal{M}}L_\text{CE}(f_{\text{MLP}}(\mathbf{x}_{j};\mathbf{W}_{\text{MLP}}), \mathbf{y}_{j})$ 
    
     $\mathcal{L} = \mathcal{L}_\text{new}+\lambda\mathcal{L}_\text{replay}$ \\
    \tcp{Update model parameters: }
    
    $\mathbf{W}_{\text{MLP}} \leftarrow \text{argmin}_{\mathbf{W}_{\text{MLP}} \in \Theta} \mathcal{L}$ 
}
Calculate importance rank $\pi_{\text{Imp}}$ by Eq.\ref{eq:attrirank_simple} 

Calculate diversity rank $\pi_{\text{Div}}$ by Eq.\ref{eq:divrank} 

Sample and update memory bank: $\mathcal{M} \leftarrow \mathcal{M} \cup \text{argtopk}\pi_{\text{Imp}} \cup \text{argtopk}\pi_{\text{Div}}$
\BlankLine
\tcc{Inference}
Initialize $f_{\text{GCN}}$ using $\mathbf{W}_{\text{MLP}}$ 

\For{$tt \leftarrow 1$ \KwTo $t$}{
    Predict on testing data: $\hat{\mathbf{Y}}_{tt}=f_{\text{GCN}}(G^{te}_{tt};\mathbf{W}_{\text{MLP}})$ 
}
} 
\end{algorithm}

\paragraph{Time Complexity}
The training process of Efficient Graph Learner is equivalent to train an $L$-layer MLP network. For simplicity, let the dimension $K$ of node attributes equal to the hidden size $D$ of MLP. The time complexity of Efficient Graph Learner is $O(L|\mathcal{V}|D^2)$, far less than the $O(L|\mathcal{V}|^2D+L|\mathcal{V}|D^2)$ of GNNs.

Regarding the sampling phase, its overhead is relatively small compared to model training as it is executed only once per task. Even so, the time complexity of importance sampling has been optimized to $O(|\mathcal{V}|D^2)$ according to the derivation in Appendix \ref{appendix:proof2}, and diversity sampling takes $O(|\mathcal{V}|\delta D)$, where $\delta$ denotes the mean degree of each node. Both complexities are linear with respect to $|\mathcal{V}|$, making them suitable for large graphs.

\paragraph{Space Complexity}
During training, since the Efficient Graph Learner does not involve graph structure, its space complexity is $O(L|\mathcal{V}|D+LD^2)$ which is more memory-efficient than other GNN-based methods $O(|\mathcal{V}|^2 + L|\mathcal{V}|D+LD^2)$. As a result, our method can allow large batch size or even whole batch training for large graphs, which will be discussed in experiments.

\section{Experiments}
In this section, we provide detailed information regarding the experiment settings, quantitative results (including both task-IL and class-IL settings), running time analysis, ablation studies, hyperparameter sensitivity analysis, and visualizations. Further information on the hyperparameter initialization, and computation resources can be found in the Appendix \ref{appendix:b}.

\begin{table}[htb]
\caption{The statistics of the node classification datasets.}
\label{tab:dataset}
\centering
\scalebox{0.8}{    
    \begin{tabular}{c|c|c|c|c}
    \toprule
         Datasets   & CoraFull & OGBN-Arxiv & Reddit & OGBN-Products \\
    \midrule
         \#Nodes &19,793 & 169,343 & 227,853 & 2,449,028\\
         \#Edges  &130,622 & 1,166,243 & 114,615,892 & 61,859,036\\
         \#Classes & 70 & 40 & 40 & 46\\
         \#Tasks  & 14 & 10 & 10 & 23\\
         \#Cls Per Task &5&4&4&2\\
    \bottomrule
    \end{tabular}
}
\end{table}
\subsection{Experimental Settings}
\paragraph{Task-IL and Class-IL Settings}
In a benchmark study conducted by CGLB \cite{cglb}, two settings for continual graph learning were introduced: task-incremental (\textbf{task-IL}) and class-incremental (\textbf{class-IL}).
In the task-IL setting, a task indicator guides the model in distinguishing between classes within each task. On the other hand, in the class-IL setting, task indicators are not provided, thus requiring the model to differentiate among all classes from both current and previous tasks.

To elucidate these settings, we utilize examples from their work. Consider a model trained on a citation network with a sequence of two tasks: {\textit{(physics, chemistry), (biology, math)}}.
In the task-IL setting, a document comes with a task indicator specifying whether it belongs to \textit{(physics, chemistry)} or \textit{(biology, math)}. Consequently, the model only needs to classify the document into one of these two task categories without distinguishing between all four classes.
Conversely, in the class-IL setting, a document can belong to any of the four classes, and the model must classify it into one of these four classes.

The class-IL setting is generally more challenging as it requires the model to handle an increasing number of classes over time. Without loss of generality, we conducted experiments on both task-IL and class-IL settings.

\paragraph{Datasets}

We conducted experiments on four node classification datasets: CoraFull~\cite{Bojchevski2017DeepGE}, Reddit~\cite{graphsage}, OGBN-Arxiv \cite{ogb}, and OGBN-Products \cite{ogb}. Following the methodology outlined by CGLB \cite{cglb}, we partitioned the label space of the original datasets into several non-intersecting segments to simulate the task/class incremental setting. For example, in the case of CoraFull, we divided the original dataset into fourteen tasks, each comprising a five-way node classification task. The detailed statistics of the datasets are presented in Table \ref{tab:dataset}.

\newcolumntype{G}{>{\columncolor{gray!25}}c}
\begin{table*}[htb]
\caption{Results on node classification task under \textbf{task-IL} setting ($\uparrow$ higher is better). Note: For joint training, it is trained on all tasks as the upper bound, therefore we do not report its AF values; For DyGRAIN, there is no open source code, so we use the results reported in their paper\cite{dygrain}; OOM indicates out of memory; The best results are marked in \textbf{bold}, and the second best results are marked \underline{underline}.}
\label{tab:node}
\centering
\scalebox{0.72}{
\begin{tabular}{l|l|cc|cc|cc|cc}
\toprule 
\multirow{2}{*}{Category}    &\multirow{2}{*}{Methods} & \multicolumn{2}{c|}{ CoraFull } & \multicolumn{2}{c|}{ OGBN-Arxiv } & \multicolumn{2}{c|}{Reddit} & \multicolumn{2}{c}{OGBN-Products}  \\
\cline{3-10} &&AA/\%$\uparrow$    &AF/\%$\uparrow$    &AA/\%$\uparrow$    &AF/\%$\uparrow$    &AA/\%$\uparrow$    &AF/\%$\uparrow$    &AA/\%$\uparrow$    &AF/\%$\uparrow$\\
\toprule 
\multirow{2}{*}{\makecell{Lower-bound \\ Upper-bound}}   &Fine-tune&39.5$\pm$1.8&-54.6$\pm$1.7  &51.5$\pm$4.6&-34.7$\pm$5.0  &57.9$\pm$3.5&-44.9$\pm$3.9      &68.7$\pm$2.0&-27.1$\pm$2.4\\
&Joint    &91.0$\pm$0.2   &N/A    &83.8$\pm$0.5  &N/A    &98.2$\pm$0.1  &N/A        &94.4$\pm$0.3&N/A\\  
\midrule
\multirow{4}{*}{CV-based}   &LwF     &56.7$\pm$5.9&-35.2$\pm$6.4  &79.2$\pm$0.7&-2.1$\pm$0.8   &72.6$\pm$6.4&-28.1$\pm$7.2      &73.8$\pm$2.0&-22.4$\pm$1.8\\       
&EWC     &66.4$\pm$3.3&-24.3$\pm$3.6  &62.3$\pm$4.3&-16.3$\pm$4.5  &87.5$\pm$4.4&-11.3$\pm$5.4      &92.3$\pm$1.4&-1.0$\pm$0.3\\    
&MAS                            &86.4$\pm$0.8&\underline{-0.2}$\pm$0.3      &\underline{81.0}$\pm$0.8&\underline{0.0}$\pm$0.0    &89.7$\pm$1.7&\textbf{0.0}$\pm$0.2    &83.6$\pm$1.0&\textbf{-0.1}$\pm$0.1\\   
&GEM     &81.1$\pm$0.8&-8.8$\pm$1.1   &67.9$\pm$0.4&-9.3$\pm$0.6   &55.6$\pm$23.3&-4.9$\pm$3.9      &78.8$\pm$15.3&-1.7$\pm$3.0\\        
\midrule
\multirow{5}{*}{Graph-based}&TWP     &86.4$\pm$0.4&-1.8$\pm$0.7   &80.5$\pm$0.8&-1.7$\pm$0.8   &87.8$\pm$5.5&-11.5$\pm$6.1      &93.1$\pm$1.5&-1.2$\pm$0.9\\         
&ER-GNN  &\underline{88.9}$\pm$0.1&\textbf{0.1}$\pm$0.3    &75.0$\pm$0.3&-8.2$\pm$0.5   &89.8$\pm$0.6&-9.2$\pm$0.8      &\underline{93.3}$\pm$0.7&\underline{-0.9}$\pm$0.1\\
&DyGRAIN  &N/A&N/A             &70.9$\pm$0.3&-4.6$\pm$0.1   &92.1$\pm$0.5&-3.5$\pm$0.1   &73.4$\pm$0.2&-3.3$\pm$0.1\\
&SSM                &\underline{88.9}$\pm$0.5&-0.8$\pm$0.1  &80.8$\pm$1.3&-2.4$\pm$0.7      &\textbf{93.7}$\pm$0.3&-4.9$\pm$0.6  &91.6$\pm$0.6&-3.6$\pm$0.2\\
&CaT &84.3$\pm$0.5&-6.1$\pm$0.8 &70.6$\pm$4.2&-14.2$\pm$2.3     &90.9$\pm$0.3&-5.5$\pm$0.5                     &OOM&OOM\\
\midrule
Ours    &\ours &\textbf{89.6}$\pm$0.1&-2.5$\pm$0.2  &   \textbf{82.1}$\pm$1.0&\textbf{0.2}$\pm$0.2    &   \underline{92.2}$\pm$0.7&\underline{-2.7}$\pm$0.8       &\textbf{93.9}$\pm$0.6&-1.2$\pm$0.3\\
\bottomrule
\end{tabular}
}
\end{table*}
\begin{table*}[htb]
    \caption{Results on node classification task under \textbf{class-IL} setting ($\uparrow$ higher is better). Note: For joint training, it is trained on all tasks as the upper bound, therefore we do not report its AF values; The results for DyGRAIN under class-IL setting are not provided in their original paper; OOM indicates out of memory; The best results are marked in \textbf{bold}, and the second best results are marked \underline{underline}.}
    \label{tab:classIL}
    \centering
    \scalebox{0.72}{
    \begin{tabular}{l|l|cc|cc|cc|cc}
    \toprule
    \multirow{2}{*}{Category}    &\multirow{2}{*}{Methods} & \multicolumn{2}{c|}{ CoraFull } & \multicolumn{2}{c|}{ OGBN-Arxiv } & \multicolumn{2}{c|}{Reddit} & \multicolumn{2}{c}{ OGBN-Products }\\
    \cline{3-10} &&AA/\%$\uparrow$    &AF/\%$\uparrow$    &AA/\%$\uparrow$    &AF/\%$\uparrow$    &AA/\%$\uparrow$    &AF/\%$\uparrow$    &AA/\%$\uparrow$    &AF/\%$\uparrow$\\
    \toprule
    \multirow{2}{*}{\makecell{Lower-bound \\ Upper-bound}}   &Fine-tune &5.5$\pm$0.0&-90.7$\pm$0.3      &8.5$\pm$0.3&-80.9$\pm$0.5  &11.6$\pm$1.3&-93.6$\pm$2.5   &4.4$\pm$1.4&-83.7$\pm$3.5\\
    &Joint   &79.5$\pm$0.1&N/A      &51.2$\pm$0.2&N/A &97.2$\pm$0.5&N/A  &76.3$\pm$0.2&N/A\\
    \midrule
    \multirow{4}{*}{CV-based}   &LwF    &5.9$\pm$0.5&-91.2$\pm$0.8      &8.8$\pm$0.2&-81.3$\pm$0.5          &11.8$\pm$0.4&-89.8$\pm$2.6     &5.4$\pm$0.1&-88.7$\pm$1.1   \\       
    &EWC                                &5.4$\pm$0.1&-91.1$\pm$3.4  &8.8$\pm$0.2&-83.1$\pm$1.4          &11.9$\pm$2.1&-95.3$\pm$1.7 &3.5$\pm$0.3&-91.8$\pm$0.9\\ 
    &MAS                                &4.3$\pm$0.1&-86.9$\pm$1.1  &8.7$\pm$0.0&-77.0$\pm$0.7          &9.1$\pm$0.2&-51.0$\pm$2.5  &\underline{19.1}$\pm$0.7&\underline{-16.2}$\pm$0.1\\   
    &GEM                                &6.4$\pm$0.6&-89.9$\pm$0.9 &8.8$\pm$0.1&-80.5$\pm$0.5          &9.2$\pm$1.3&\underline{-11.2}$\pm$4.8           &7.7$\pm$3.0&\textbf{-15.3}$\pm$2.6\\  
    \midrule
    \multirow{4}{*}{Graph-based}    &TWP     &7.5$\pm$0.1&-86.3$\pm$0.5      &8.5$\pm$0.1&-81.4$\pm$0.5  &13.4$\pm$1.3&-93.6$\pm$1.4     &3.3$\pm$0.6&-94.8$\pm$0.3\\
    &ER-GNN                                  &5.1$\pm$0.2&-90.7$\pm$0.2      &17.3$\pm$0.8&-69.6$\pm$1.1 &62.0$\pm$4.5&-39.4$\pm$5.1     &6.8$\pm$1.6&-90.2$\pm$2.2\\
    &SSM                                     &\textbf{76.2}$\pm$1.2&\textbf{6.0}$\pm$0.1  &\underline{33.0}$\pm$0.4&\textbf{21.5}$\pm$0.4      &\underline{74.8}$\pm$1.7&\textbf{-7.1}$\pm$0.7          &OOM&OOM\\
    &CaT                    &14.8$\pm$5.2&-81.3$\pm$3.7  &32.1$\pm$0.3&-48.8$\pm$0.5    &74.1$\pm$2.2&-20.8$\pm$1.2         &OOM&OOM\\
    \midrule
    Ours    &E-CGL   &\underline{68.2}$\pm$0.2&\underline{-17.8}$\pm$0.2&\textbf{33.7}$\pm$0.2&\underline{-41.0}$\pm$0.1   &\textbf{78.6}$\pm$1.0&-17.3$\pm$1.2     &\textbf{56.3}$\pm$1.0&-31.0$\pm$1.1\\
    
    \bottomrule
    \end{tabular}
    }
\end{table*}

\paragraph{Baselines}
We compared E-CGL against nine state-of-the-art continual learning techniques. These include four conventional methods (LwF \cite{lwf}, EWC \cite{ewc}, MAS \cite{mas}, and GEM \cite{gem}) originally designed for CV tasks, along with five methods specifically tailored for graph tasks (TWP \cite{twp}, ER-GNN \cite{ergnn}, DyGRAIN \cite{dygrain}, SSM \cite{ssm}, and CaT \cite{cat}). Additionally, we included fine-tuning and joint training techniques as performance lower-bound and upper-bound, respectively.

\paragraph{Metrics}
Following the definitions provided in \cite{cglb}, we employed the performance matrix $\mathbf{M}^p\in\mathcal{R}^{T\times T}$ to depict the dynamics of overall performance, where $\mathbf{M}^p_{i,j}$ denotes the accuracy performance at task $j$ after the model has been trained on task $i$. Based on this performance matrix, we computed the average accuracy (AA) and average forgetting (AF) as defined in \cite{gem}: ${\frac{\sum^i_{j=1}\mathbf{M}^p_{i,j}}{i}|i=1,...,T}$ for AA, and ${\frac{\sum^{i-1}_{j=1}\mathbf{M}^p_{i,j}-\mathbf{M}^p_{j,j}}{i-1}|i=2,...,T}$ for AF. It's important to note that in the continual learning scenario, AF is typically negative, indicating the occurrence of catastrophic forgetting.

\begin{table*}[htb]
    \caption{Running time (train/inference) comparison of different methods under task-IL setting. The E-GCL (GCN) is the version that we use GCN as encoder. The \textit{Improv.} indicates the relative improvement our E-CGL compared with E-CGL (GCN). We report both training and inference time, and the time unit is milliseconds (ms).}
    \label{tab:time}
    \centering
    \scalebox{0.85}{
    \begin{tabular}{l|ccccG}
    \toprule
    Methods   & CoraFull     &  OGBN-Arxiv   &  Reddit &OGBN-Products &\textit{Avg. Time} \\
    \toprule
    TWP                 &34.25/105.71          &37.33/23.49          &354.68/1104.26        &3217.46/2894.05    &910.93/1031.88\\
    ER-GNN              &56.51/107.67          &47.17/22.64          &421.83/1075.20        &4419.80/4385.23    &1236.33/1397.69\\
    SSM                 &52.68/109.89     &49.63/27.65    &491.99/1220.88   &3976.74/4351.92    &1142.76/1427.59\\
    CaT     &176.24/115.81   &152.13/23.15  &1719.85/1120.00    &OOM    &N/A\\
    \midrule
    E-CGL(GCN)          &28.94/110.79       &43.18/27.39    &1078.31/1032.93    &5205.96/4912.76    &1589.10/1520.97\\
    E-CGL               &\textbf{15.69}/\textbf{99.24}          &\textbf{25.61}/\textbf{17.95}          &\textbf{177.13}/\textbf{574.13}       &\textbf{183.04}/\textbf{552.72}      &\textbf{100.37}/\textbf{311.01}\\
    \textit{Improv.}    &1.94$\times$/1.12$\times$   &1.69$\times$/1.53$\times$   &6.08$\times$/1.80$\times$   &28.44$\times$/8.89$\times$  &15.83$\times$/4.89$\times$\\
    \bottomrule
    \end{tabular}
    }
\end{table*}

\subsection{Task-IL Results and Analysis}
\label{sec:4.2}

The results for task-IL node classification are presented in Table \ref{tab:node}. Our proposed E-CGL method demonstrates notable performance among the competing methods. It achieves the highest average accuracy (AA) on the CoraFull, OGBN-Arxiv, and OGBN-Products datasets, and the second-highest average accuracy on Reddit. On the OGBN-Arxiv dataset, E-CGL surpasses the second-best result by a margin of 1.1\%, closely approaching the performance upper bound of joint training. Notably, E-CGL only employs simple MLP for training, while still outperforms its GNNs competitors, which highlights the effectiveness of our method. Additionally, E-CGL exhibits an average forgetting of -1.1\% across four datasets, indicating its ability to effectively mitigate catastrophic forgetting.

Moreover, we compared the performance of different methods with the lower bound (fine-tune) and upper bound (joint train) as benchmarks. Nearly all methods fall between these bounds, indicating varying degrees of improvement over fine-tuning. However, most methods still suffer from catastrophic forgetting, as indicated by negative average forgetting (AF) values. Additionally, it is worth mentioning that graph-specific continual learning methods, tend to rank higher compared to other CV-based methods. This observation emphasizes the existence of a representation gap between graph data and other Euclidean data, underscoring the necessity of tailored approaches for graph-related tasks.

\begin{table*}[htb]
\caption{Ablation study results of E-CGL under task-IL setting, where we removed the proposed \textit{importance sampler} and \textit{diversity sampler} for replay, respectively. And \textit{-MLP,+GCN} means we employed GCN instead of MLP for training.}
\label{tab:ablation}
\centering
\scalebox{0.8}{
\begin{tabular}{l|cc|cc|cc|cc}
\toprule
\multirow{2}{*}{Methods} & \multicolumn{2}{c|}{ CoraFull } & \multicolumn{2}{c|}{ OGBN-Arxiv } & \multicolumn{2}{c|}{ Reddit } & \multicolumn{2}{c}{ OGBN-Products }\\
\cline{2-9}& AA / \% $\uparrow$ & AF / \% $\uparrow$ & AA / \% $\uparrow$ & AF / \% $\uparrow$ & AA / \% $\uparrow$ & AF / \% $\uparrow$ & AA / \% $\uparrow$ & AF / \% $\uparrow$\\
\toprule
E-CGL                &\textbf{89.6}$\pm$0.1&-2.5$\pm$0.2  &82.1$\pm$1.0&0.2$\pm$0.1   &92.2$\pm$0.7&-2.7$\pm$0.8  &93.9$\pm$0.6&-1.2$\pm$0.3\\
-importance sampler  &88.3$\pm$0.2&1.5$\pm$0.2   &80.0$\pm$0.6&-3.2$\pm$0.5  &88.9$\pm$2.0&-1.4$\pm$2.1  &93.7$\pm$0.4&-0.3$\pm$0.1\\
-diversity sampler   &89.3$\pm$0.4&-2.0$\pm$0.1           &81.5$\pm$0.7&-1.0$\pm$0.1    &90.6$\pm$1.8&-3.1$\pm$1.5  &93.2$\pm$1.2&-2.5$\pm$0.9\\
-MLP,+GCN            &88.8$\pm$0.7&-2.9$\pm$0.9 &\textbf{82.7}$\pm$0.3&0.8$\pm$0.6   &\textbf{95.8}$\pm$1.9&-1.6$\pm$1.2  &\textbf{94.0}$\pm$0.6&0.0$\pm$0.4\\
\bottomrule
\end{tabular}
}
\end{table*}

\subsection{Class-IL Results and Analysis}

The experimental results under the class-IL setting are depicted in Table \ref{tab:classIL}. Firstly, it's evident that all statistics in Table \ref{tab:classIL} are notably worse compared to the task-IL results in Table \ref{tab:node}. This disparity arises due to the increased difficulty of the class-IL prediction setting, which involves a much larger label space. For example, on the CoraFull dataset, the classifier trained under the task-IL setting only predicts within 5 possible labels, while under the class-IL setting, it must discern among all 70 classes.
Consequently, the outcomes of four CV-based traditional methods cannot fit in this setting and have poor performances similar to the fine-tuning.

Secondly, the graph-regularization method TWP~\cite{twp} performs poorly and is even indistinguishable to the fine-tuning model. This may be attributed to strong distribution discrepancies across different tasks, where the regularization approach struggles to maintain performance on previous tasks without access to previous data. This further underscores the effectiveness of utilizing replay-based methods for continual graph learning.

Thirdly, the replay-based methods (including ER-GNN~\cite{ergnn}, SSM~\cite{ssm}, CaT~\cite{cat}, and our E-CGL) show notable performance improvements. However, ER-GNN's performance is highly dependent on the dataset, achieving acceptable results on OGBN-Arxiv and Reddit but showing poor performance on CoraFull and OGBN-Products. Additionally, SSM and CaT fail to operate effectively on the large OGBN-Products dataset and report out-of-memory error. In contrast, E-CGL demonstrates more stable performance and satisfactory cost across all datasets, yielding acceptable results. This can be attributed to a more efficient sampling strategy that covers a wider feature space.

Finally, joint training demonstrates dominant performance across all datasets, indicating that catastrophic forgetting remains a significant factor in the class-IL setting. While retraining all previous data can effectively mitigate the forgetting problem, it comes with unacceptable computational costs.

\begin{figure*}[htb]
    \centering
    \subfigure[The y-axis is average accuracy (AA).]{\includegraphics[scale=0.37]{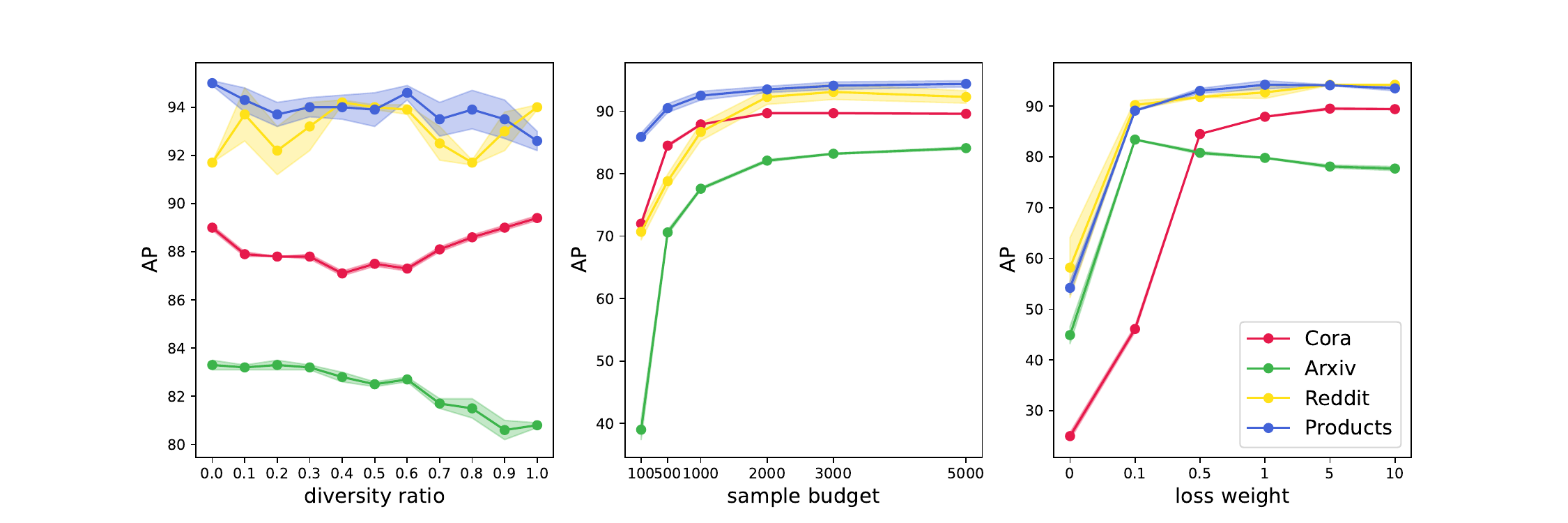}}\\
    \subfigure[The y-axis is average forgetting (AF).]{\includegraphics[scale=0.37]{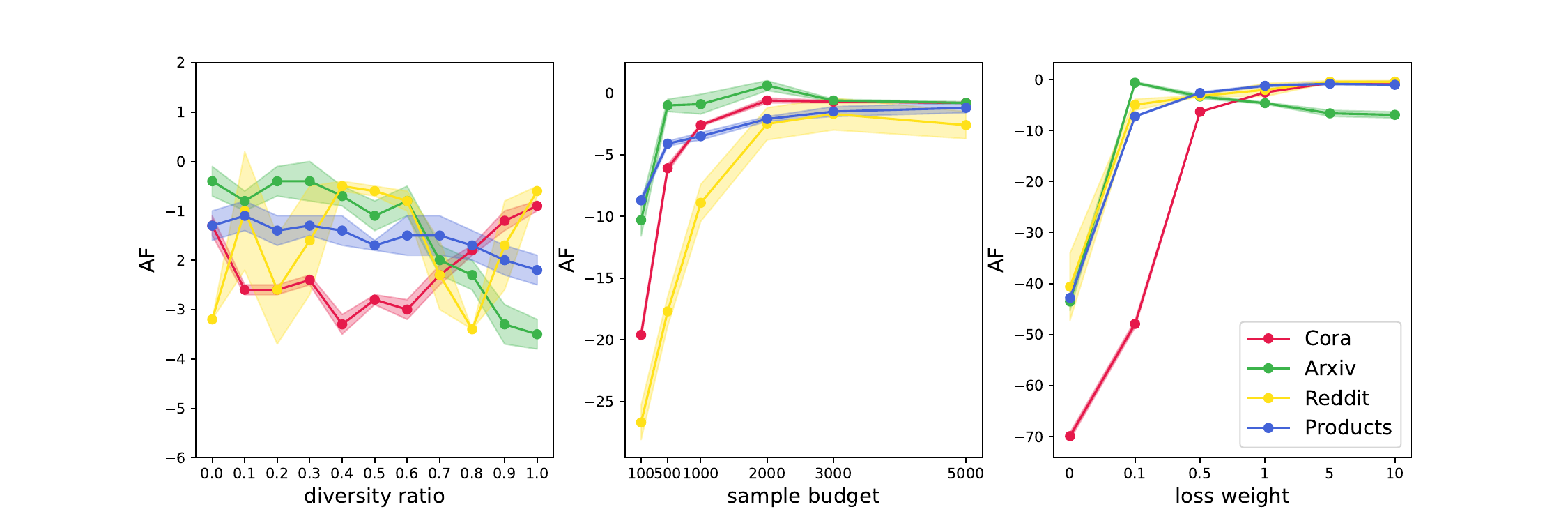}}
    \vspace{-1em}
    \caption{Parameters sensitivity analysis on E-CGL, the shallow shades are variances. Left: diversity sampling ratio. Middle: sampling budget for Graph Dependent Replay. Right: loss weight $\lambda$.}
    \label{fig:sensitivity_af}
\end{figure*}

\begin{figure*}[htb]
    \centering
    \includegraphics[width=0.9\linewidth, height=0.28\textwidth]{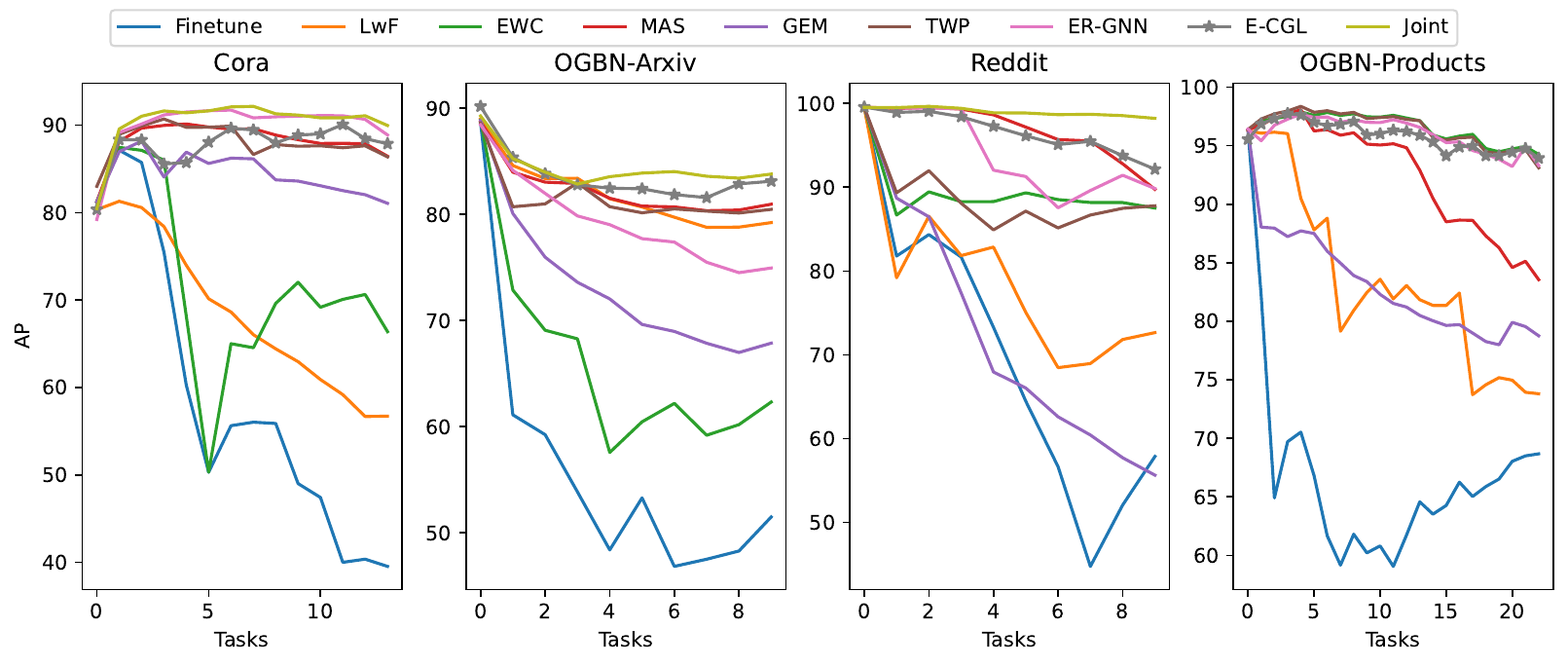}
    \vspace{-1em}
    \caption{Visualization: Learning curves of AA over task sequences. Note: The curve for joint training on OGBN-Products is unavailable due to resource limitations.}
    \label{fig:learning_curve}
\end{figure*}

\subsection{Running Time Analysis}

To evaluate the efficiency of E-CGL, we conducted a running time analysis comparing our method with other graph continual learning methods under the task-IL setting. The results, presented in Table \ref{tab:time}, show the average time required for each training and inference epoch.

We observed that E-CGL significantly reduces both training and inference time compared to its GCN version. On average, E-CGL achieves a speedup of 15.83 times during training and 4.89 times during inference. The magnitude of improvement becomes more significant as the dataset size increases. For example, the improvement on the OGBN-Products dataset reaches 28.44 times during training and 8.89 times during inference, respectively. Additionally, E-CGL comprehensively outperforms other continual graph learning methods in terms of both training and inference time.

These efficiency improvements can be attributed to the design of the Efficient Graph Learner, where E-CGL eliminates the need for time-consuming message passing during training and allows for direct training on large graphs without batching.
The reduced training time of E-CGL makes it a practical and efficient solution for continual graph learning tasks, enabling faster experimentation, model iteration, and deployment.

\begin{figure*}[htb]
    \centering
    \subfigure[CoraFull]{
        \includegraphics[scale=0.22]{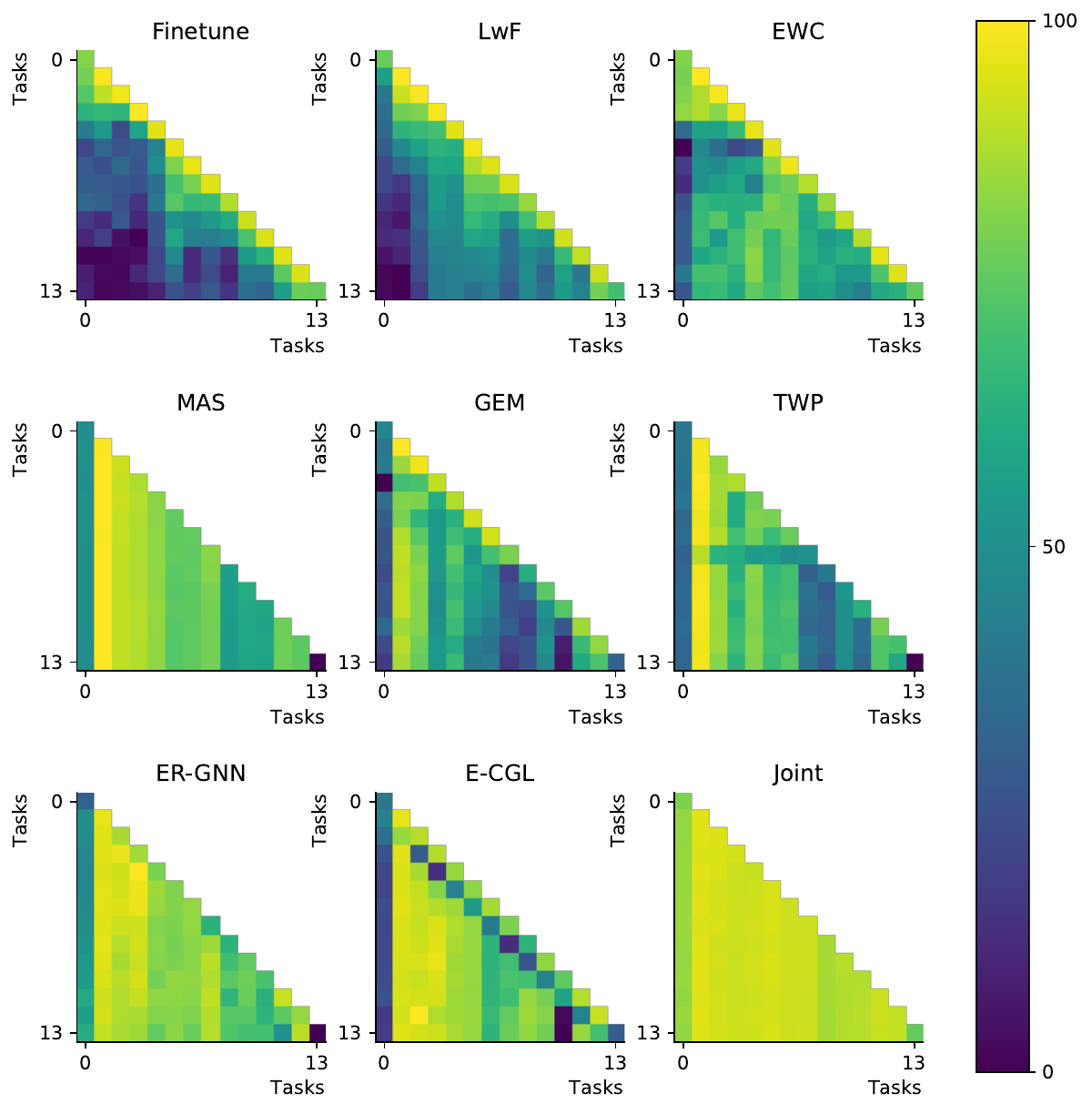}
    }
    \subfigure[OGBN-Arxiv]{
        \includegraphics[scale=0.22]{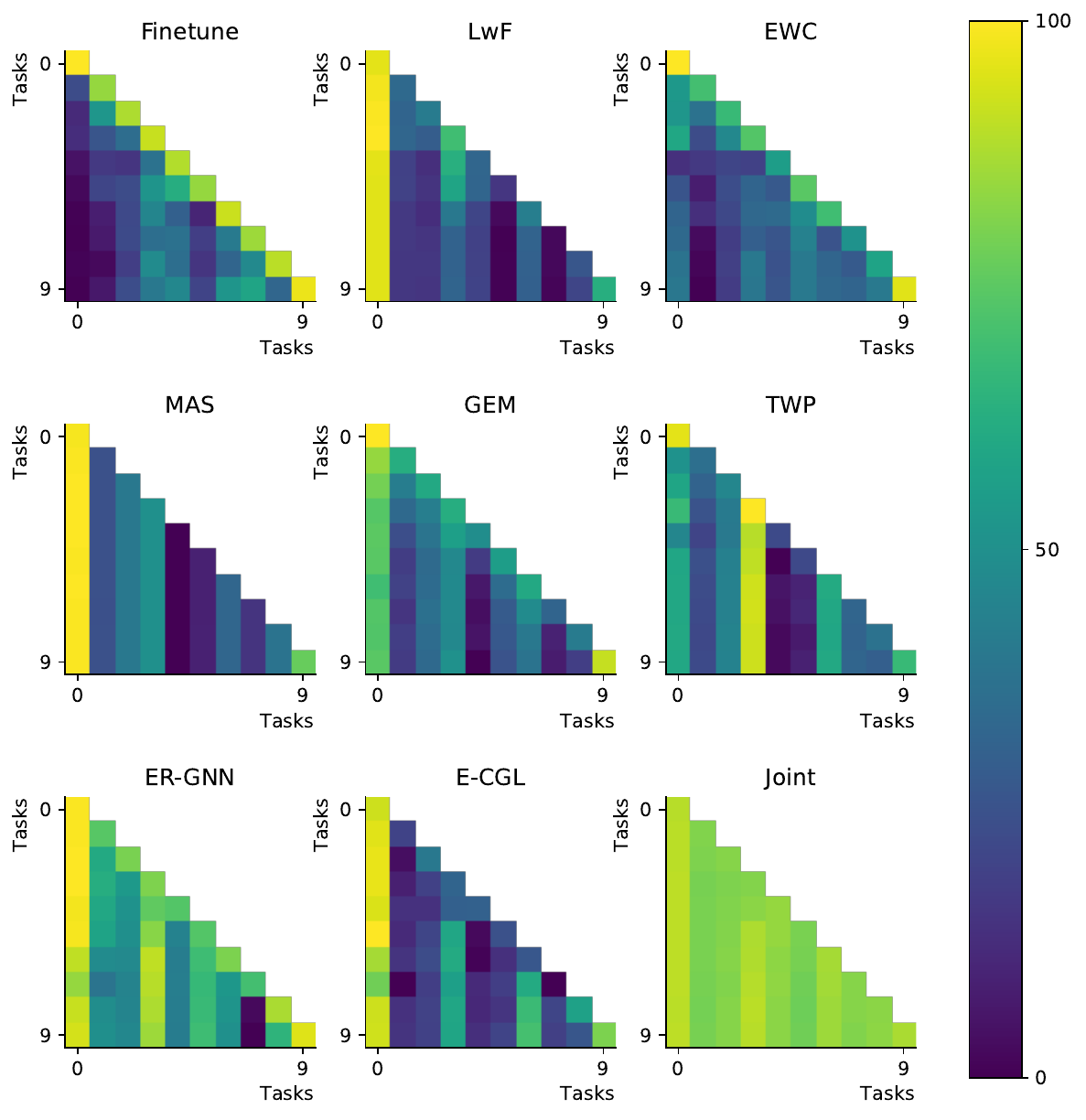}
    }
    \subfigure[Reddit]{
        \includegraphics[scale=0.22]{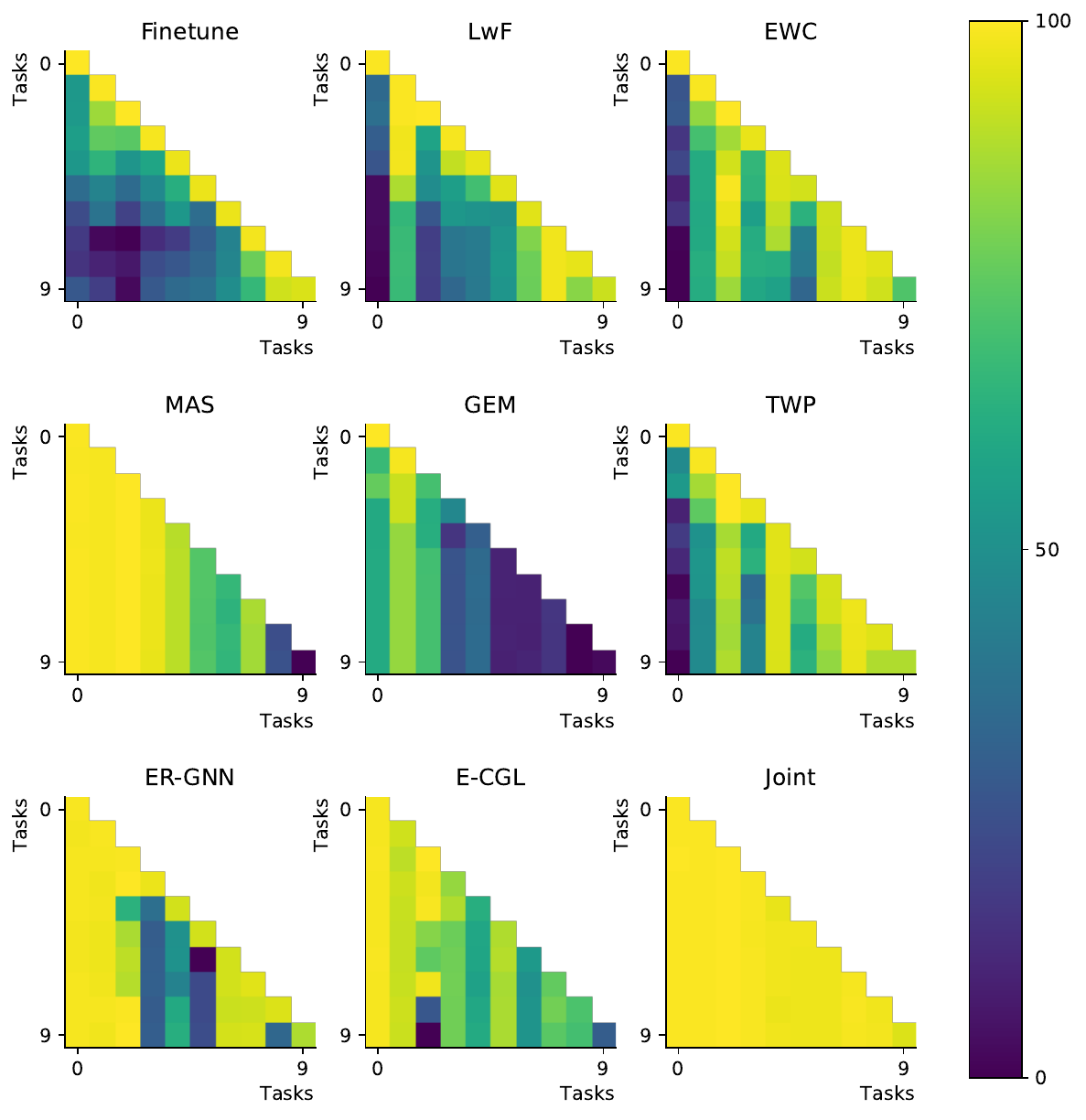}
    }
    \subfigure[OGBN-Products]{
        \includegraphics[scale=0.22]{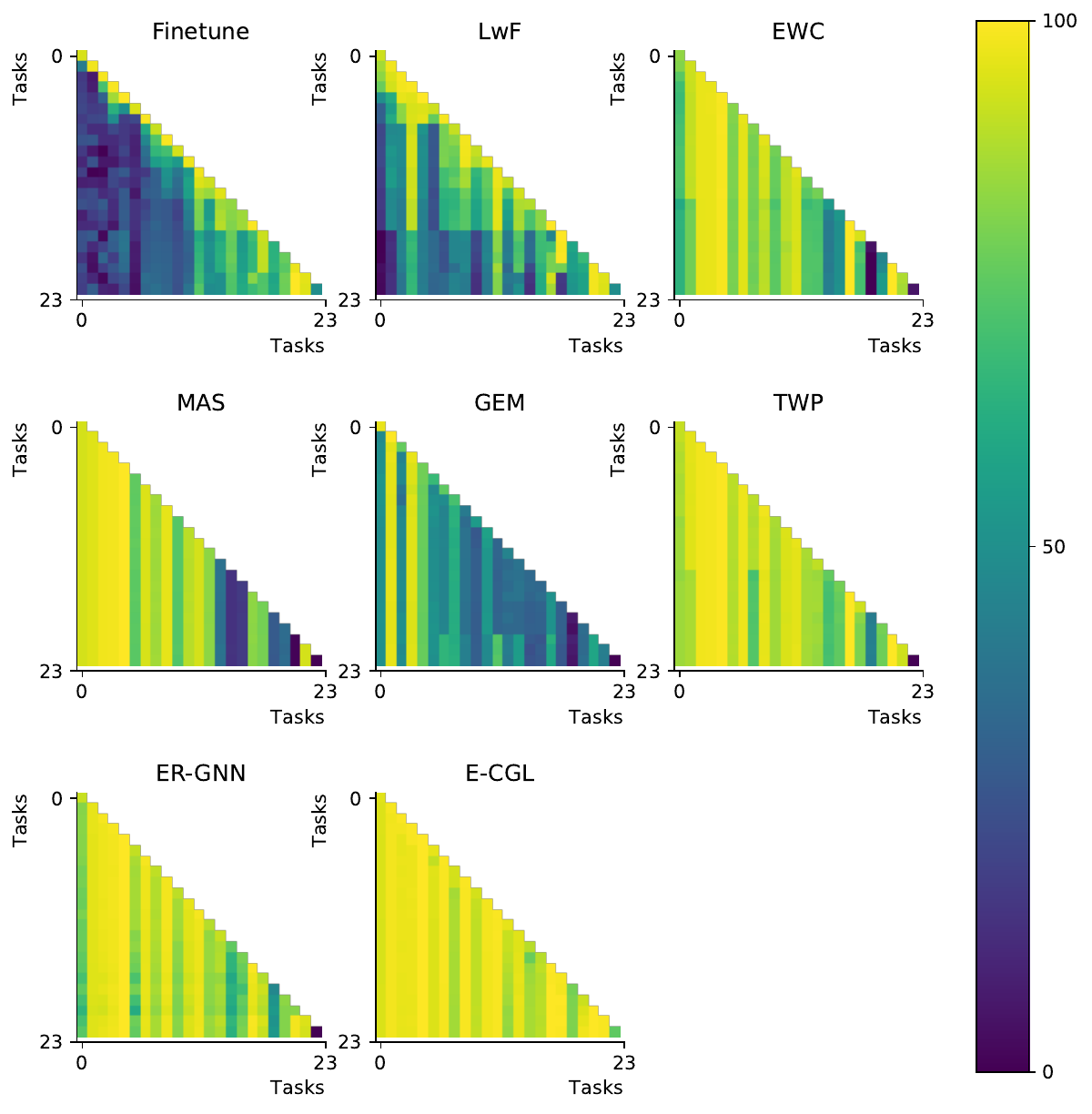}
    }
    \caption{Visualization: Performance matrices on CoraFull, OGBN-Arxiv, Reddit, and OGBN-Products.}
    \label{fig:performance_matrix}
\end{figure*}



\subsection{Ablation Study}

To assess the effectiveness of different components in E-CGL, we conducted ablation studies under the task-IL setting. Two key components were evaluated: the replay-based sampling strategy and the choice of training encoder (MLP vs. GCN). The results are presented in Table \ref{tab:ablation}.

Firstly, we examined the impact of removing the \textit{importance} and \textit{diversity} samplers, respectively, while keeping the same sampling budget for the other. The results indicate that the performance of E-CGL decreases in terms of AA when either sampler is removed, underscoring the importance of both components of the proposed replay-based sampling strategy.
Secondly, it is notable that in most cases, removing the \textit{importance sampler} resulted in a larger performance drop compared to removing the \textit{diversity} sampler. 
Additionally, the magnitude of the difference also reflects the dataset's difficulty, as will be discussed in the subsequent visualization section. On simpler datasets such as Products, there is minimal difference between the full version and the version with removed samplers, whereas on relatively more challenging datasets like Arxiv and Reddit, the choice of sampling strategy has a more significant impact on the results.

Furthermore, we replaced the Efficient Graph Learner (MLP-based encoder) with GCN as the training encoder, as shown in the last row of Table \ref{tab:ablation}. While it is not surprising that the GCN version outperforms the MLP version in most cases, given the importance of graph topological information for graph training, it is worth noting that the performance gap is not substantial. This suggests that despite a minor performance degradation, our E-CGL still achieves acceptable results while offering a significant training speedup compared to GCN-based methods.

\subsection{Parameter Sensitivity}

We conducted parameter sensitivity analysis on three factors: \textit{1) diversity sampling ratio}, \textit{2) sampling budget for $\mathcal{M}$}, and \textit{3) loss weight $\lambda$}. The average accuracy (AA) and the average forgetting (AF) results are depicted in Figure \ref{fig:sensitivity_af}.

Firstly, we observed that the diversity sampling ratio has a minimal impact on the results of continual graph learning. Both AA and AF exhibit slight fluctuations within a narrow range as the diversity ratio varies. The differing curve trends also indicate that the effect of different sampling strategies depends on specific datasets, thus highlighting the necessity of a combined strategy.

Secondly, the performance of E-CGL consistently improves with an increase in the sampling budget. This is aligned with expectations, as a larger budget enables the replay of more nodes, effectively enhancing model performance. However, it's important to note that the budget cannot be infinitely expanded due to storage limitations and concerns regarding training efficiency. Eventually, when the sampling budget becomes extremely large, the replay-based method converges to joint training.

Lastly, the model's performance initially increases and then decreases (more noticeably on OGBN-Arxiv) as the loss weight $\lambda$ is raised. This pattern aligns with intuition. In our main experiment, we simply set $\lambda$ to 1.

\subsection{Visualization}
\subsubsection{Learning Curve}
\label{sec:4.5}

We also visualized the learning curves of certain methods in Figure \ref{fig:learning_curve} to provide further insight into the training process over the task sequence.

The learning curves demonstrate a decreasing trend for all methods, indicating the presence of catastrophic forgetting. The rate of decline in the learning curves provides an indication of the difficulty of the datasets. Arxiv and Reddit, which exhibit steeper declines in AA values, are relatively challenging datasets compared to the simpler Products dataset.

An important observation is that the replay-based approaches, including ER-GNN and our proposed E-CGL, consistently outperform other methods. This finding suggests that the interdependencies in graph data are strong and need to be explicitly maintained to mitigate catastrophic forgetting. The replay-based methods, by leveraging past samples during training, are able to retain important information and sustain performance on previously learned tasks.

\subsubsection{Performance Matrix}

We have also generated performance matrices to visualize the performance of several baseline methods and E-CGL under task-IL setting, as depicted in Figure \ref{fig:performance_matrix}. Similar to the analysis conducted in the previous section \ref{sec:4.2}, most methods fall between the lower bound of fine-tuning and the upper bound of joint training. Generally, E-CGL demonstrates higher values, and graph-specific continual learning techniques show better overall performance.

One interesting finding in the performance matrix is related to the diagonal entries, which represent the model's ability to adapt to new tasks. It can be observed that in certain cases, some regularization methods (\textit{e.g.,} TWP on CoraFull, GEM on Products) exhibit weaker adaptation ability. We speculate that these methods impose constraints on model parameter updates, which can preserve the model's performance on previous tasks but limit its ability to adapt to new tasks.

The overall value range of the performance matrix also directly reflects the difficulty of each dataset. It is noticeable that most methods have darker (\textit{i.e.,} lower) values on OGBN-Arxiv compared to OGBN-Products. This observation aligns with the findings discussed in section \ref{sec:4.5}.
\section{Conclusion}

In this paper, we have tackled two challenging obstacles in continual graph learning: the interdependencies in graph data and the efficiency concerns associated with growing graphs. To address these challenges, we have introduced an efficient continual graph learner (E-CGL) that utilizes a graph-specific sampling strategy for replay and an MLP encoder for efficient training. Our extensive empirical results demonstrate the effectiveness of our method in terms of both performance and efficiency.

Despite the growing interest in continual graph learning, there are still some limitations that need to be addressed. For instance, there is a need to explore methods for protecting historical data privacy during the replay phase and actively forgetting unneeded stale knowledge for model adaptation. Furthermore, further investigation into handling heterogeneous graphs and addressing graph classification tasks is warranted.
We hope that E-CGL serves as a step forward in continual graph learning and inspires future research to explore broader applications and develop more advanced techniques in continual graph learning domain.







\bibliography{ref}
\appendix
\section{Theoretical Proofs}
\subsection{Proof for Equation \ref{eq:surrogate}}
\label{appendix:proof1}

Based on the definition of $Q$ and $r$ in Equation \ref{eq:Q} and \ref{eq:r} respectively, we can derive that~\cite{attrirank}:
\begin{equation}
\begin{aligned}
(Q r)_i & =\sum_{j \in \mathcal{V}} \frac{s_{i j}}{\sum_{k \in \mathcal{V}} s_{k j}} r_j \\
& =\sum_{j \in \mathcal{V}} \frac{s_{i j}}{\sum_{k \in \mathcal{V}} s_{k j}} \frac{\sum_{k \in \mathcal{V}} s_{j k}}{z} \\
& =\frac{1}{z} \sum_{j \in \mathcal{V}} s_{i j} \\
& =1 \cdot r_i.
\end{aligned}
\end{equation}

Therefore we have $\mathbf{1}\cdot r = Qr$.

\subsection{Simplify r Using Taylor Expansion}
\label{appendix:proof2}
Consider the $r_i$ in Equation \ref{eq:r} being unnormalized: $\hat{r}_i=\sum_{j \in \mathcal{V}}s(i,j)$ and with the Radial Basis Function $s(i,j)=e^{-\gamma||\mathbf{x}_i-\mathbf{x}_j||_2^2}$ as similarity, it can be simplified using Taylor expansion~\cite{attrirank}: 
\begin{equation}
\begin{aligned}
\hat{r}_i  =&\sum_{j \in \mathcal{V}} e^{-\gamma\left\|x_i-x_j\right\|_2^2}\\
=&\sum_{j \in \mathcal{V}} e^{-\gamma\left(\left\|x_i\right\|_2^2+\left\|x_j\right\|_2^2-2 x_i^T x_j\right)} \\
\approx& e^{-\gamma\left\|x_i\right\|_2^2} \sum_{j \in \mathcal{V}} e^{-\gamma\left\|x_j\right\|_2^2}\left(1+2 \gamma x_i^T x_j+\frac{1}{2}\left(2 \gamma x_i^T x_j\right)^2\right) \\
=&e^{-\gamma\left\|x_i\right\|_2^2}\left[\sum_{j \in \mathcal{V}} e^{-\gamma\left\|x_j\right\|_2^2}+x_i^T\left(2 \gamma \sum_{j \in \mathcal{V}} e^{-\gamma\left\|x_j\right\|_2^2} x_j\right)\right. \\
&\left.+x_i^T\left(2 \gamma^2 \sum_{j \in \mathcal{V}} e^{-\gamma\left\|x_j\right\|_2^2} x_j x_j^T\right) x_i\right] \\
\equiv & w_i\left[a+x_i^T b+x_i^T C x_i\right] ,
\end{aligned}
\end{equation}
where $w_i = e^{-\gamma\left\|x_i\right\|_2^2}$, $a=\sum_{j \in \mathcal{V}} e^{-\gamma\left\|x_j\right\|_2^2}$ , $b=2 \gamma \sum_{j \in \mathcal{V}} e^{-\gamma\left\|x_j\right\|_2^2} x_j$, and $c=2 \gamma^2 \sum_{j \in \mathcal{V}} e^{-\gamma\left\|x_j\right\|_2^2} x_j x_j^T$ can be pre-calculated and all of them requires an $O(|\mathcal{V}|D^2)$ time complexity. Considering $|\mathcal{V}|\gg D$ in most cases, such cost is acceptable for large graphs.

\section{Implementation Details}
\label{appendix:b}
\subsection{Running Environment}
The experiments were conducted on a machine with NVIDIA 3090 GPU (24GB memory). 
The E-CGL model and other baselines were implemented using Python 3.9.16\footnote{https://www.python.org/downloads/release/python-3916/}, PyTorch 1.12.1\footnote{https://pytorch.org/get-started/previous-versions/}, CUDA 11.3\footnote{https://developer.nvidia.com/cuda-11.3.0-download-archive}, and DGL 0.9.1\footnote{https://www.dgl.ai/}.
The code was developed based on the benchmark CGLB\cite{cglb}.

\subsection{Model Configurations}
For a fair comparison, a two-layer GCN~\cite{gcn} with a hidden dimension of 256 is used as the backbone for all compared methods. 
Unless otherwise specified, the same training configurations, including optimizer, learning rate, weight decay, and training epochs, are used for all baseline methods.
Specifically, a batch size of 8000 is used when batching is necessary. Adam is employed as the optimizer with a learning rate of 0.005, and the weight decay is set to $5\times10^{-4}$. Each task is trained for 200 epochs.
The reported mean and standard deviations of AP and AF are based on five independent runs with random seeds ranging from 0 to 4.

\subsection{Hyperparameters}
Comprehensive hyperparameters specific to each method are provided in Table \ref{tab:hyperparameters}, and the reported results are based on the best outcomes obtained through grid search on these hyperparameters.
To reproduce the results of E-CGL, set the sampling budget for Graph Dependent Replay as 1000 for CoraFull, 3000 for OGBN-Arxiv, 5000 for Reddit and OGBN-Products. Among all sampled nodes, 25\% are selected using diversity sampling and 75\% are selected using importance sampling. The loss weight $\lambda$ is set to 1.

\begin{table}[ht]
    \caption{Hyperparameters list for compared methods.}
    \label{tab:hyperparameters}
    \centering
    \scalebox{0.9}{
    \begin{tabular}{c|c}
    \toprule
         Methods&  Hyperparameters\\
    \midrule
        LwF     &$\lambda_{\text{dist}}: \{1.0, 10.0\}$, $\text{T}: \{2.0, 20.0\}$ \\ 
        EWC     &$\text{memory strength}: \{10000.0\}$\\
        MAS     &$\text{memory strength}: \{10000.0\}$\\
        GEM     &$\text{memory strength}: \{0.5\}$, $\text{memory nums}:\{100\}$\\
        TWP     &$\lambda_{\text{l}}: \{10000.0\}$, $\lambda_{\text{t}}: \{10000.0\}$, $\beta: \{0.01\}$\\
        ER-GNN  &$\text{sample budget}:\{500,1000,5000\}$, $d:\{0.5\}$, $\text{sampler}: \{\text{MF}, \text{CM}\}$\\
        DyGRAIN &$\text{results are imported from their original paper}$\\
        SSM     &$\text{c\_node budget}: \{100\}$, $\text{neighbor budget}: \{[0, 0]\}$, $\lambda: 1$\\
        CaT     &$\text{sample budget}:\{100, 1000, 3000, 5000\}$\\
        E-CGL   &$\text{sample budget}:\{1000, 3000, 5000\}$, $\text{diversity ratio}: \{0.1,0.25\}$, $\lambda:\{1.0\}$\\
    \bottomrule
    \end{tabular}
    }
\end{table}

\end{document}